\documentclass[acmtog,nonacm,screen]{acmart} 

\usepackage{wrapfig}
\usepackage{makecell}
\usepackage{enumitem}
\usepackage{soul}
\usepackage{xcolor}
\usepackage{xspace}
\usepackage{colortbl}
\definecolor{colorTrd}{rgb}{0.95,0.95,0.65}
\definecolor{colorSnd}{rgb}{1, 0.85, 0.7}
\definecolor{colorFst}{rgb}{1, 0.7, 0.7}

\usepackage{amsfonts,pifont,bm,mathtools,nicefrac}
\usepackage{multirow,textcomp}
\usepackage[capitalize]{cleveref}  
\crefname{section}{Sec.}{Secs.}
\Crefname{section}{Section}{Sections}
\crefname{table}{Tab.}{Tabs.}
\Crefname{table}{Table}{Tables}
\crefname{figure}{Fig.}{Figs.}
\Crefname{figure}{Figure}{Figures}
\crefname{equation}{Eq.}{Eqs.}
\Crefname{equation}{Equation}{Equations}
\usepackage[ruled]{algorithm2e} 

\SetAlFnt{\small}
\SetAlCapFnt{\small}
\SetAlCapNameFnt{\small}
\SetAlCapHSkip{0pt}

\begin{document}
\title{StableIntrinsic: Detail-preserving One-step Diffusion Model for Multi-view Material Estimation}

\author{Xiuchao Wu}
\authornote{Joint first authors}
\email{wuxiuchao@zju.edu.cn}
\affiliation{%
  \institution{State Key Lab of CAD\&CG, Zhejiang University}
  \country{China}
}
\affiliation{%
  \institution{Alibaba Group}
  \country{China}
}
\author{Pengfei Zhu}
\authornotemark[1]
\email{pfzhu@smail.nju.edu.cn}
\affiliation{%
  \institution{State Key Lab for Novel Software Technology, Nanjing University}
  \country{China}
}
\affiliation{%
  \institution{Alibaba Group}
  \country{China}
}

\author{Jiangjing Lyu}
\authornote{Project leader}
\email{jiangjing.ljj@taobao.com}
\affiliation{%
  \institution{Alibaba Group}
  \country{China}
}

\author{Xinguo Liu}
\email{xgliu@cad.zju.edu.cn}
\affiliation{%
 \institution{State Key Laboratory of CAD\&CG, Zhejiang University}
 \country{China}
}

\author{Jie Guo}
\email{guojie@nju.edu.cn}
\affiliation{%
 \institution{State Key Lab for Novel Software Technology, Nanjing University}
 \country{China}
}

\author{Yanwen Guo}
\email{ywguo@nju.edu.cn}
\affiliation{%
 \institution{State Key Lab for Novel Software Technology, Nanjing University}
 \country{China}
}

\author{Weiwei Xu}
\email{xww@cad.zju.edu.cn}
\affiliation{%
  \institution{State Key Lab of CAD\&CG, Zhejiang University}
  \country{China}
}

\author{Chengfei Lyu}
\authornote{Corresponding author}
\email{chengfei.lcf@taobao.com}
\affiliation{%
  \institution{Alibaba Group}
  \country{China}
}





\newcommand{\point}{\mathbf{x}}

\newcommand{\viMLP}{f^s_\theta}
\newcommand{\vdMLP}{f^r_\theta}

\newcommand{\camPos}{\mathbf{o}}
\newcommand{\camRot}{\mathbf{R}}
\newcommand{\ray}{\mathbf{r}}
\newcommand{\inputRay}{\hat{\ray}}
\newcommand{\inputRaySet}{\hat{\mathcal{R}}}
\newcommand{\image}{I}
\newcommand{\imageSet}{\mathcal{\image}}
\newcommand{\inputImage}{\hat{\image}}
\newcommand{\inputImageSet}{\hat{\mathcal{\image}}}
\newcommand{\pose}{T}
\newcommand{\poseSet}{\mathcal{\pose}}
\newcommand{\inputPose}{\hat{\pose}}
\newcommand{\inputPoseSet}{\hat{\mathcal{\pose}}}
\newcommand{\inputColor}{\hat{\mathbf{c}}}
\newcommand{\avgColor}{\bar{\mathbf{c}}}
\newcommand{\radianceField}{\Theta}
\newcommand{\viColor}{\mathbf{c}_d}
\newcommand{\vdColor}{\mathbf{c}_s}

\newcommand{\density}{\sigma}
\newcommand{\viDensity}{\density^s}
\newcommand{\vdDensity}{\density^r}

\newcommand{\tile}{k}
\newcommand{\tileSet}{\mathcal{K}}
\newcommand{\tileCandidates}{\tileSet^c}

\newcommand{\lossColor}{\mathcal{L}_{c}}
\newcommand{\lossWarp}{\mathcal{L}_{\text{WARP}}}
\newcommand{\lossDecom}{\mathcal{L}_{\text{DECOM}}}
\newcommand{\lossDepth}{\mathcal{L}_{DEPTH}}
\newcommand{\lossSmooth}{\mathcal{L}_{SMOOTH}}
\newcommand{\lossADMM}{\mathcal{L}_{ADMM}}
\newcommand{\lossReg}{\mathcal{L}_{\text{REG}}}
\newcommand{\lossConsistency}{\mathcal{L}_{CON}}
\newcommand{\lossPerview}{\mathcal{L}_{SELF}}
\newcommand{\lossInterpolation}{\mathcal{L}_{BLEND}}

\newcommand{\lambdaColor}{\lambda_{c}}
\newcommand{\lambdaSSIM}{\lambda_{\text{SSIM}}}
\newcommand{\lambdaVGG}{\lambda_{\text{VGG}}}
\newcommand{\lambdaSurfaceColor}{\lambda_{s}}
\newcommand{\lambdaReg}{\lambda_{\text{REG}}}


\newcommand{\sceneStreet}{\emph{Street}\xspace}
\newcommand{\sceneShadyPath}{\emph{Shady Path}\xspace}
\newcommand{\sceneCommunity}{\emph{Community}\xspace}
\newcommand{\scenePark}{\emph{Park}\xspace}
\newcommand{\sceneRubble}{\emph{Rubble}\xspace}
\newcommand{\scenePolytech}{\emph{Polytech}\xspace}
\newcommand{\sceneBar}{\emph{Bar}\xspace}
\newcommand{\sceneSofa}{\emph{Sofa}\xspace}
\newcommand{\sceneCoffeeShop}{\emph{Coffee Shop}\xspace}
\newcommand{\sceneLivingRoomInria}{\emph{Living Room from Inria}~\cite{philip2021free}\xspace}
\newcommand{\sceneLivingRoomXu}{\emph{Living Room from Xu et al.}~\shortcite{XuSIG21}\xspace}

\newcommand{\ourDataset}{\emph{LGDM}\xspace}
\newcommand{\sceneBlueCar}{\emph{Blue Car}\xspace}
\newcommand{\sceneRedCar}{\emph{Red Car}\xspace}
\newcommand{\sceneNatatorium}{\emph{Natatorium}\xspace}
\newcommand{\sceneGlassBust}{\emph{Glass Bust}\xspace}
\newcommand{\sceneSkyscraper}{\emph{Skyscraper}\xspace}
\newcommand{\sceneMall}{\emph{Mall}\xspace}
\newcommand{\sceneBull}{\emph{Bull}\xspace}
\newcommand{\sceneSculpture}{\emph{Sculpture}\xspace}

\newcommand{\flux}{\Phi}
\newcommand{\normal}{\mathbf{n}}
\newcommand{\pos}{\mathbf{x}}
\newcommand{\xpos}{\mathbf{x}}
\newcommand{\ypos}{\mathbf{y}}
\newcommand{\direction}{\boldsymbol{\omega}}
\newcommand{\dirin}{\direction_\text{i}}
\newcommand{\dirout}{\direction_\text{o}}
\newcommand{\hemisphere}{\mathcal{H}^2}
\newcommand{\withnorm}{[\normal]}
\newcommand{\withgeom}{[\normal_g]}
\newcommand{\withshading}{[\normal_s]}
\newcommand{\withboth}{[\normal_g, \normal_s]}
\newcommand{\inout}{\dirin \rightarrow \dirout}
\newcommand{\outin}{\dirout \rightarrow \dirin}
\newcommand{\emitted}{L_\text{e}}
\newcommand{\importance}{W_\text{e}}
\newcommand{\freeflight}{p_\text{ff}}
\newcommand{\im}{\mathrm{i}}
\newcommand{\tnear}{{t_\text{n}}}
\newcommand{\tfar}{{t_\text{f}}}
\newcommand{\network}{F_{\boldsymbol\theta}}
\newcommand{\rgbL}{L_{\text{RGB}}}
\newcommand{\irL}{L_{\text{IR}}}
\newcommand{\tofL}{L_{\text{ToF}}}
\newcommand{\rgbhatL}{\hat{L}_{\text{RGB}}}
\newcommand{\irhatL}{\hat{L}_{\text{IR}}}
\newcommand{\tofhatL}{\hat{L}_{\text{ToF}}}
\newcommand{\Iref}{I_{\text{s}}}
\newcommand{\timet}{\tau}
\newcommand{\latent}{\mathbf{z}}
\newcommand{\static}{\text{stat}}
\newcommand{\dynamic}{\text{dyn}}

\newcommand{\hashFeature}{\mathbf{f}}
\newcommand{\hashGrid}{\Phi}

\newcommand{\abs}[1]{\left\lvert#1\right\rvert}
\newcommand{\norm}[1]{\left\lVert#1\right\rVert}

\newcommand{\degree}{\textdegree\xspace}

\newcommand{\etal}{et al.}
\newcommand{\eg}{e.g.}
\newcommand{\ie}{i.e.}

\newcommand{\final}{\textup{final}}
\newcommand{\gt}{\textup{gt}}

\newcommand{\red}[1]{#1}

\newcommand{\weiwei}[1]{{\color{red} {\bf Weiwei:} #1}}
\newcommand{\xww}[1]{\weiwei{#1}}

\newcommand{\jiamin}[1]{{\color{purple} {\bf Jiamin:} #1}}

\newcommand{\jt}[1]{{\color{magenta} {\bf JT:} #1}}

\definecolor{darkgreen}{rgb}{0,0.55,0}
\newcommand{\xchao}[1]{ \noindent {\color{darkgreen} {\bf xchao:} {#1}} }
\newcommand{\pf}[1]{ \noindent {\color{blue} {\bf pfz:} {#1}} }

\newcommand{\qixing}[1]{{\color{mycolor2} {\bf Qixing:} #1}}

\newcommand{\zihan}[1]{ \noindent {\color{violet} {\bf Zihan:} {#1}} }

\newcommand{\tocite}[1]{\textcolor{red}{[TO CITE]}}
\newcommand{\change}[1]{\textcolor{red}{#1}}
\newcommand{\method}{TODO\xspace}
\newcommand{\supp}{\textit{Supplementary Material}\xspace}
\newcommand{\yj}[1]{{\color{blue} [Yujun: #1]}}

\newcommand{\evan}[1]{{\color[rgb]{0.1,0.5,0.5}{\bf Evan: #1}}}
\newcommand{\todo}[1]{{\color[rgb]{0.75,0.0,0.5}{TODO: #1}}}

\begin{abstract}

Recovering material information from images has been extensively studied in computer graphics and vision. Recent works in material estimation leverage diffusion model showing promising results. However, these diffusion-based methods adopt a multi-step denoising strategy, which is time-consuming for each estimation. Such stochastic inference also conflicts with the deterministic material estimation task, leading to a high variance estimated results. In this paper, we introduce StableIntrinsic, a one-step diffusion model for multi-view material estimation that can produce high-quality material parameters with low variance. To address the overly-smoothing problem in one-step diffusion, 
StableIntrinsic applies losses in pixel space, with each loss designed based on the properties of the material. 
Additionally, StableIntrinsic introduces a Detail Injection Network (DIN) to eliminate the detail loss caused by VAE encoding, while further enhancing the sharpness of material prediction results. The experimental results indicate that our method surpasses the current state-of-the-art techniques by achieving a $9.9\%$ improvement in the Peak Signal-to-Noise Ratio (PSNR) of albedo, and by reducing the Mean Square Error (MSE) for metallic and roughness by $44.4\%$ and $60.0\%$, respectively.

\end{abstract}

%
%


%
%

\begin{teaserfigure}
\centering
  \includegraphics[width=1.0\textwidth]{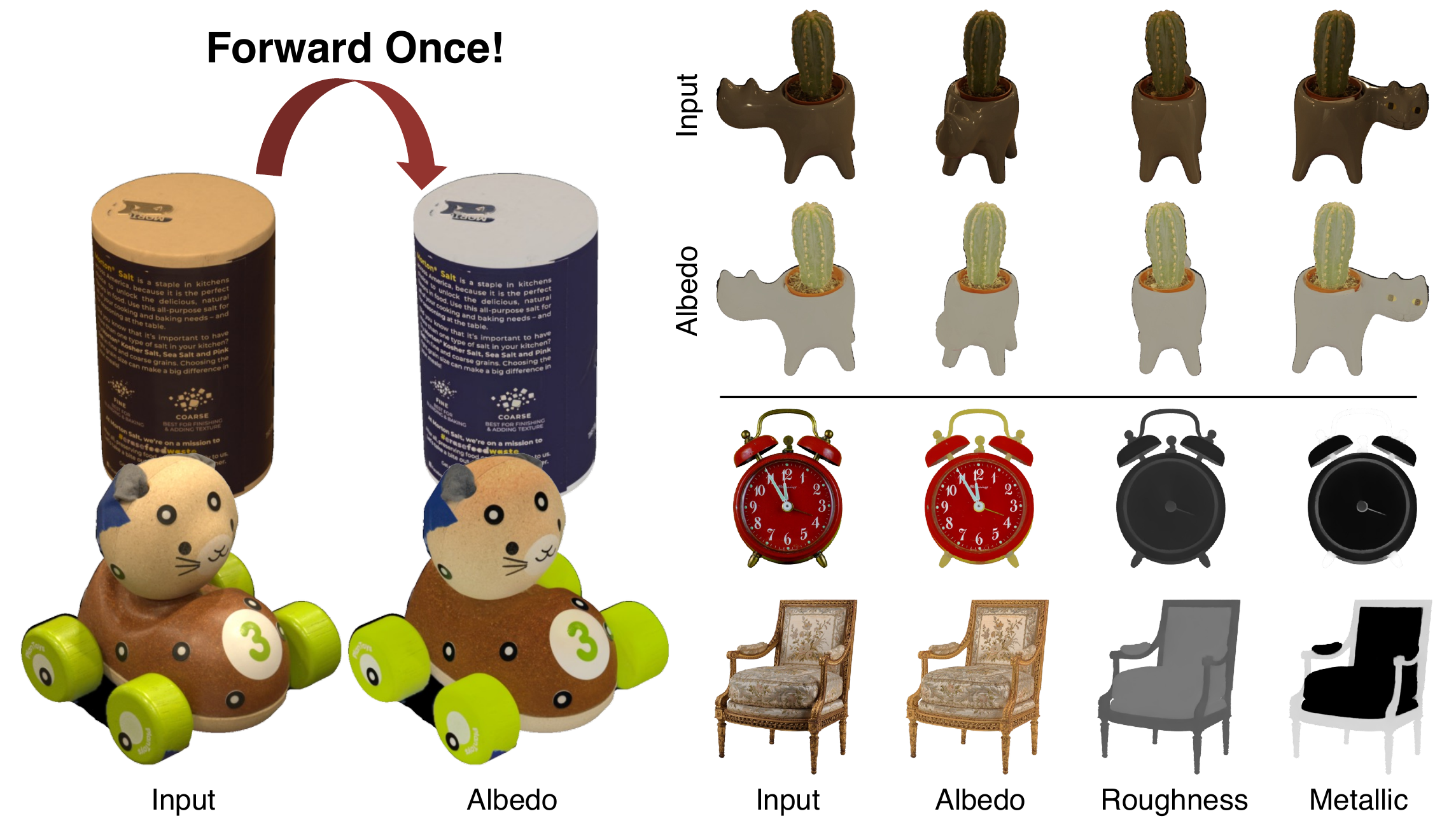}
  \caption{\textbf{Estimated results of StableIntrinsic.} \textbf{Left:} StableIntrinsic estimates material information from RGB images in a single step, producing results with preserved details; \textbf{Top right:} StableIntrinsic supports multi-view material estimation and can output globally consistent material maps; \textbf{Bottom right:} StableIntrinsic can accurately estimate material parameters for different types of objects.}
  \label{fig:teaser}
\end{teaserfigure}


\maketitle

\section{Introduction}\label{sec:intro}
Intrinsic image decomposition is a long-standing challenge aiming to retrieve the surface reflectance property and the effects of illumination separately. The former, known as material estimation, is a fundamental but ill-posed problem~\cite{grosse2009ground}.
In light of the remarkable success of diffusion-based generative models~\cite{ho2020denoising, rombach2022high} in dense perception tasks, recent works~\cite{zeng2024rgb, chen2024intrinsicanything,li2024idarb} integrate these models into material estimation to address the issue of overly-smoothing and detail-lacking results of regression-based methods.
This is achieved through dozens to hundreds of forward passes using the UNet architecture to denoise Gaussian noise~\cite{rombach2022high,song2020denoising}, which typically requires several to tens of seconds for inference on a general graphics card. 

\begin{figure}
\centering
\includegraphics[width=\linewidth]{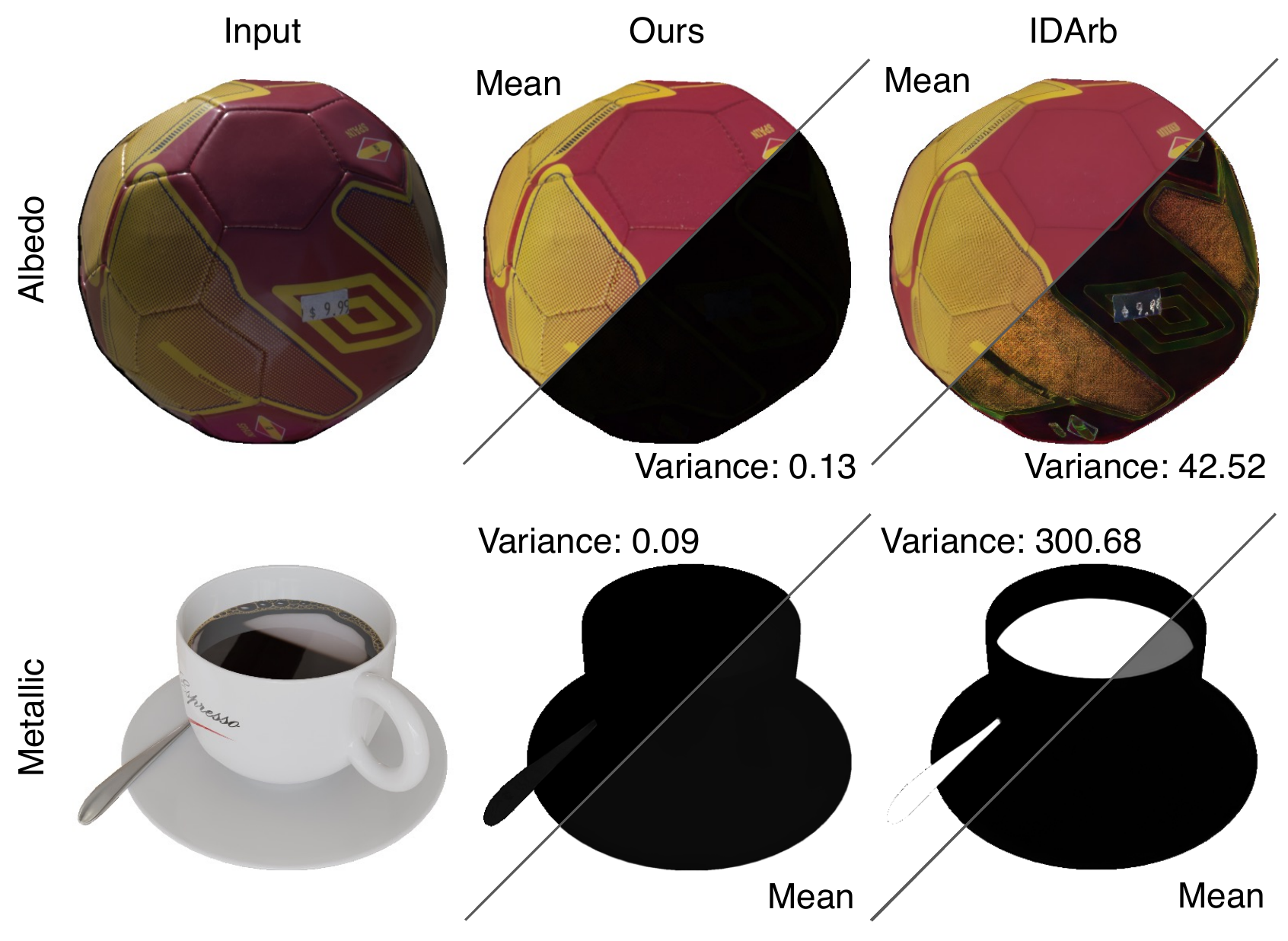}
\caption{\textbf{Low-variance material estimation of StableIntrinsic.} We use multiple different random noise to inference and  visualize the mean and variance of the predicted material maps for StableIntrinsic and SOTA method IDArb~\cite{li2024idarb}.}
\label{fig:var} 
\end{figure}

In addition to inefficiency, these approaches also suffer from inherent stochasticity.
As shown in Fig.~\ref{fig:var}, the results generated by this kind of approach often exhibit significant variance. 
For albedo, as high-frequency areas are more easily affected by the input Gaussian noise, these areas have high variance across different samples, leading to various generated textures.
Although these textures might seem perceptually correct, the misalignment with actual labels results in substantial pixel-level errors.
For roughness and metallicity, as the images are captured under uncontrollable illumination, the ambiguity in disentangling reflection causes multiple possible explanations of the material. 
Although the correct one usually has the highest probability of being produced, the randomness of input Gaussian noise might still affect the model to produce those low-probability incorrect explanations. This is often observed for glossy regions (Fig.~\ref{fig:var} bottom).
Such significant variance makes it unreliable and limits its applicability for deterministic tasks. Overall, it is necessary to develop an efficient and low-variance model for material estimation.

\begin{figure}[t]
\centering
\includegraphics[width=\linewidth]{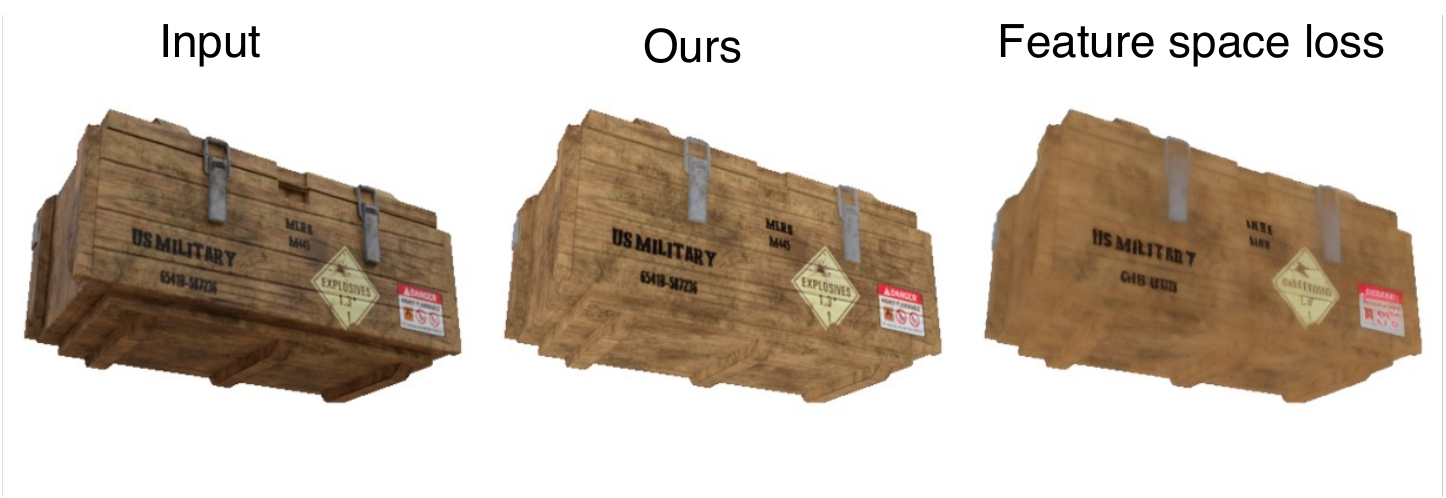}
\caption{\textbf{Overly-smoothing problem of applying loss in feature space.}}
\label{fig:feature_loss} 
\end{figure}

To this end, we propose StableIntrinsic, a diffusion-based model that can estimate material parameters from multi-view RGB images in a single step (Fig.~\ref{fig:teaser}). 
Instead of forwarding multiple times for denoising, StableIntrinsic only needs to forward once, resulting in a significant acceleration of the estimation process. 
It also naturally lowers the variance of the output results (Fig.~\ref{fig:var}). 
The key challenge of one-step diffusion model is the blurry output. To overcome this, rather than optimizing in feature space, which is ineffective for producing higher quality results (Fig.~\ref{fig:feature_loss}), we choose to optimize StableIntrinsic in pixel space~\cite{xu2024matters}.
This effectively improves the output results, especially in high-frequency areas. 
Additionally, images with rich detailed textures fail to be well preserved in estimated results as the VAE breaks the structure information of them after encoding.
We thus introduce a Detail Injection Network (DIN) to further correct the results and enhance the sharpness.
In general, StableIntrinsic has three contributions:
\begin{enumerate}[label=\textbullet,leftmargin=14pt]
\item To the best of our knowledge, StableIntrinsic is the first one-step diffusion model for material estimation, which significantly accelerates inference and enhances estimation stability.
\item We address the overly-smoothing problem of one-step denoising by optimizing StableIntrinsic in pixel space and introducing a Detail Injection Network (DIN).
\item Extensive experiments show that StableIntrinsic outperforms state-of-the-art methods, showing high quality estimated albedo with rich details, and accurate metallic and roughness.  
\end{enumerate}

\begin{figure*}
\centering
\includegraphics[width=\linewidth]{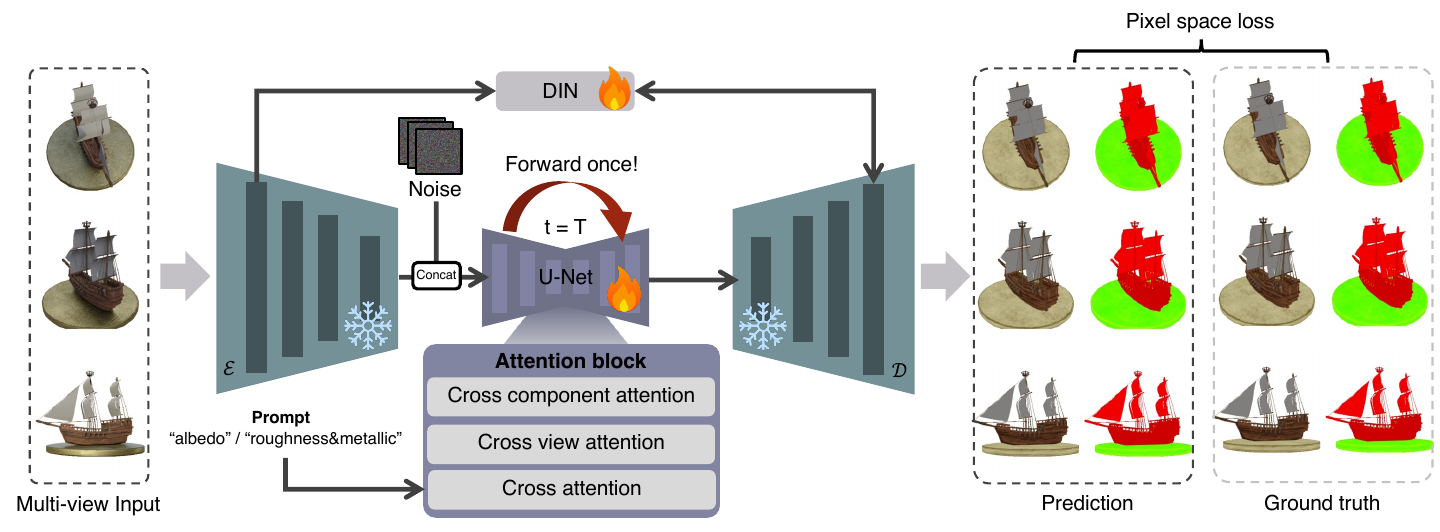}
\caption{\textbf{An overview of our method.} StableIntrinsic predicts material parameters of multi-view RGB images in one-step. We optimize our model in pixel space to enhance the acquisition of fine-grained details. Based on the property of each material, we use different loss for albedo $\bf A$ and roughness\&metallic $\bf RM$. To compensate for the quality degradation caused by VAE encoding, we introduce a Detail Injection Network (DIN) to further improve the output results.}
\label{fig:pipeline} 
\end{figure*}

\section{Related Works}\label{sec:related}

\subsection{Learning-based Material Estimation}

The material estimation task aims to disentangle the surface material properties and environment illumination from images~\cite{grosse2009ground,garces2012intrinsic,chen2013a,barron2013intrinsic,barron2015shape},
which is discussed in detail in Garces et al.~\shortcite{garces2022a}.
Recent advancements in deep learning have led to significant improvements in this field~\cite{Azinovic_2019_CVPR,philip2021free,wang2021learning,li2020inverse,zhu2022learning}, surpassing previous methods. 
They use feed-forward models to decompose images, but due to the ambiguities of material and illumination, they often produce blurry results lacking high-frequency details~\cite{kocsis2024iid}.

Generative models have attracted widespread attention and discussion in multiple fields~\cite{rombach2022high}. Leveraging prior learning from a massive number of high-fidelity images, diffusion-based models have demonstrated strong capabilities across various vision tasks~\cite{he2024lotus,zeng2024dilightnet,fu2024geowizard,ke2023marigold,ye2024stablenormal,liang2025diffusionrenderer,xu2024matters}. Several studies have used these models as pre-trained foundations, fine-tuning them for intrinsic decomposition~\cite{luo2024intrinsicdiffusion,kocsis2024iid,zeng2024rgb,chen2024intrinsicanything,li2024idarb,huang2024materialanything,Zhang2024MaPa,he2025neural}. Although the results that are more consistent with visual perception are achieved compared to regression-based approaches, these models have limitations. Single-image methods suffer from multiview inconsistency~\cite{kocsis2024iid,zeng2024rgb,chen2024intrinsicanything}. IDArb~\cite{li2024idarb} enhances multiview consistency with additional attention blocks but struggles with incorrect texture structures disrupted by the outermost auto-encoder and tends to oversimplify complex real-world objects. Moreover, these models often require dozens to hundreds of forward passes of the neural network, resulting in slow and costly inference, limiting their applicability to downstream tasks.

\subsection{One-step Diffusion-based Estimation}
Trivially rescheduling the noises into fewer steps or a single-step denoising often results in low-quality generation~\cite{song2020denoising}. Therefore, accelerating the sampling process has been a topic of extensive discussion, exploring methods such as distillation models~\cite{meng2022distillation,salimans2022progressive,yin2024onestep}, Rectified Flow~\cite{liu2022flow,liu2024instaflow}, Consistency Models~\cite{song2023consistency}, and Shortcut Models~\cite{frans2025one}.

When utilizing diffusion-based models in vision tasks, the strong correspondence between conditional images and ground-truth labels facilitates reconstruction, yielding clean results in single-step denoising scenarios. Various studies~\cite{xu2024matters,he2024lotus,ye2024stablenormal,martingarcia2024diffusione2eft} have shown that single-step forward delivers comparable results in dense perception tasks. Recognizing the instability due to stochastic nature of diffusion models, these studies propose deterministic approaches that exclude noisy inputs. Nevertheless, they still struggle to recover high-frequency details. To overcome this issue, GenPercept~\cite{xu2024matters} further substitutes the original decoder with a task-specific one and employs pixel-wise losses to enhance results. Lotus~\cite{he2024lotus} applies a reconstruction task as regularization, whereas StableNormal~\cite{ye2024stablenormal} utilizes a two-stage denoising framework with semantic-guided refinement.

Material estimation, a special case of dense perception tasks, necessitates detail preservation, highlight-shadow removal, and recovery of semantic-consistent material parameters. Building on prior works, our method adjusts sampling and regularization strategies to address this complex task, achieving superior performance.

\section{Method}\label{sec:method}

StableIntrinsic is a diffusion-based model, which aims to estimate high-quality PRB material (i.e. albedo $\bf A$, roughness $\bf R$ and metallicity $\bf M$) parameters from multi-view RGB images efficiently. To this end, we make the denoising process a single step to accelerate the estimation process, which also reduces the variance in the predicted material maps (Fig.~\ref{fig:var}).
To address the overly-smoothing problem of one-step estimation, we choose to optimize StableIntrinsic with pixel space losses (Sec.~\ref{subsec:pixel-loss}). Finally, we introduce a detail-injection module to further improve the output results (Sec.~\ref{subsec:detial_refine}). An overview of our method is shown in Fig.~\ref{fig:pipeline}.

\subsection{Preliminary}
\label{subsec:one-step}
Given a random noise ${\bf z}_T \sim {\mathcal N}({\bf 0, I})$, the latent diffusion model~\cite{rombach2022high} sequentially denoises it into a target data ${\bf z}_0$ by predicting noise $\epsilon_t$ at each time step $t$. The noise is typically output by a neural network $\mu_\theta$, which can be optimized with loss for $\bf v$-parameterization~\cite{salimans2022progressive}: 
\begin{align}
L_\theta =  \mathbb{E}_{\epsilon \sim {\mathcal N}({\bf 0, I}), {\bf Z_c}, t}||{\bf v} - \mu_\theta({\bf z}_t, t, {\bf z}_c)||^2_2,
\label{equ:denoise}
\end{align}
\begin{align}
{\bf v} = \sqrt{\bar{\alpha}_t}\epsilon - \sqrt{1 - \bar{\alpha}_t}{\bf z}_0,
\label{equ:vprediction}
\end{align}
where ${\bf z}_c = \mathcal{E}({\bf I})$ is the latent feature of a condition image $\bf I$ encoded by a pre-trained VAE \cite{kingma2013auto} encoder $\mathcal{E}$. $\bar{\alpha}_t$ is the noise schedule that controls the proportion of noise. ${\bf z}_t$ is sampled by adding noise to ${\bf z}_0$:
\begin{align}
{\bf z}_t = \sqrt{\bar{\alpha}_t}{\bf z}_0 + \sqrt{1 - \bar{\alpha}_t}\epsilon, \epsilon \sim {\mathcal N}({\bf 0, I}), t \in \{0,1,...,T\}.
\label{equ:addnoise}
\end{align}

The one-step denoising aims to predict ${\bf z}_0$ from ${\bf z}_T$ directly, which is a pure Gaussian noise (${\bf z}_T = \epsilon$) as $\bar {\alpha}_T = 0$. Therefore, Equ.~\ref{equ:denoise} can be reformulated as:
\begin{align}
L_\theta =  \mathbb{E}_{\epsilon \sim {\mathcal N}({\bf 0, I}), {\bf z}_c, t=T}||-{\bf z}_0 - \mu_\theta(\epsilon, T, {\bf z}_c)||^2_2.
\label{equ:one-step-denoise}
\end{align}
In this case, the output of $\mu_\theta$ is constrained to predict the negative ${\bf z}_0$ (Equ.~\ref{equ:vprediction}). As a result, instead of optimizing across different time steps, one-step denoising only needs to be optimized at time step $T$.

\begin{figure}[t]
\centering
\includegraphics[width=\linewidth]{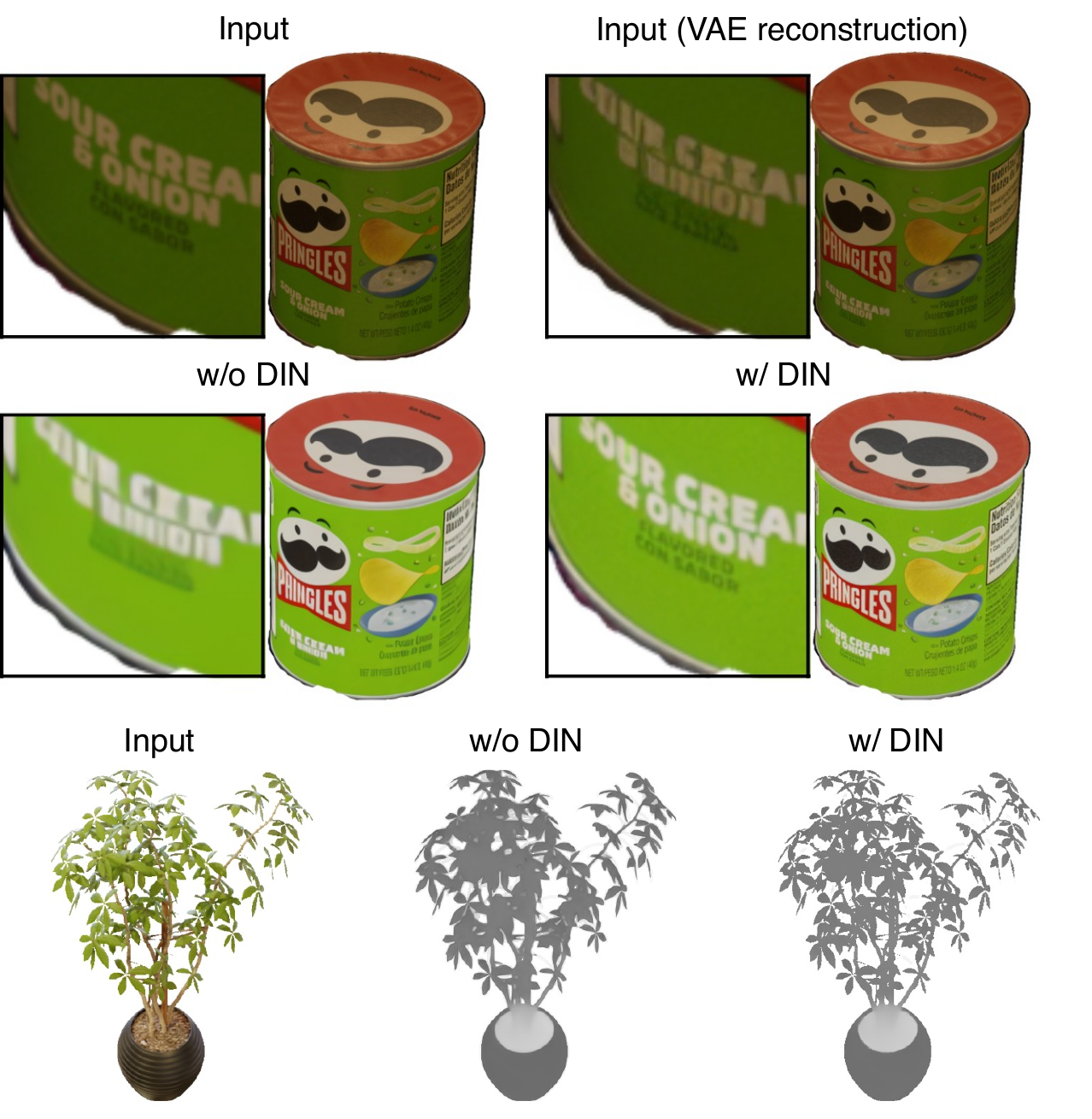}
\caption{\textbf{DIN effectively injects details.}
\textbf{Top:} VAE fails to reconstruct input condition images with rich details; \textbf{Mid:} DIN can successfully inject details to predicted albedo map while avoiding introducing reflections from the lid of this can; \textbf{Bottom:} DIN enhances the sharpness of predicted roughness.}
\label{fig:din_ablation} 
\end{figure}

\subsection{One-step material estimation}
\label{subsec:pixel-loss}
Unfortunately, simply denoising in a single step with a diffusion model is almost unable to produce details, showing overly-smoothing prediction results. To address this problem, we optimize our model in pixel space, which is similar to GenPercept~\cite{xu2024matters} and generally has two benefits:
\begin{enumerate}[leftmargin=14pt, itemsep=2pt]
\item 
Since the resolution of output images is 8 times than that of latent features, our loss is calculated at a higher resolution, which can help $\mu_\theta$ produce results with more details (Fig.~\ref{fig:feature_loss}); 
\item
Based on the properties of albedo, roughness and metallicity, we can choose different types of losses for them to achieve higher-quality results (Fig.~\ref{fig:gm}).
\end{enumerate}

Therefore, the optimization of our one-step material estimation model can be formulated as:
\begin{align}
L_\theta =  \mathbb{E}_{\epsilon \sim {\mathcal N}({\bf 0, I}), {\bf z}_c, t=T}||-{\bf K} - \mathcal{D}(\mu_\theta(\epsilon {\bf z}_c, T))||^2_2,
\label{equ:pixel-loss}
\end{align}
where $\mathcal{D}$ is a pre-trained VAE~\cite{kingma2013auto} decoder corresponding to the encoder $\mathcal{E}$.
$\bf K \in \{A, RM\}$ is the target material, including albedo ${\bf A} \in \mathbb{R}^{H \times W \times 3}$ and roughness\&metallic ${\bf RM} \in \mathbb{R}^{H \times W \times 3}$, where ${\bf RM} = [{\bf R}^{H \times W \times 1}, {\bf M}^{H \times W \times 1}, {\bf 0}^{H \times W \times 1}]$.  The input condition ${\bf z}_c$ is concatenated with $\epsilon$ before feeding to $\mu_\theta$.

During training, we use an MSE loss to supervise albedo $\bf \hat{A}$ and roughness and metallic $\bf \hat{RM}$. In addition, we use a gradient matching loss~\cite{MegaDepthLi18} for $\bf \hat{RM}$ to improve the sharpness of their boundaries:
\begin{align}
    L_{GM}(\mathbf{RM},\mathbf{\hat{RM}})=\frac{1}{N}(||\nabla_x(\mathbf{\hat{RM}}-\mathbf{RM})||^1_1+||\nabla_y(\mathbf{\hat{RM}}-\mathbf{RM})||^1_1),
\end{align}
where $N$ is the total number of pixels, and $\nabla_x$/$\nabla_y$ denotes the gradient in x/y direction. Compared to the MSE loss, which calculates the loss for each pixel equally, $L_{GM}$ is calculated based on neighboring pixels and focuses on penalizing areas with unmatched gradients (e.g., blurred RM boundaries or incorrectly baked textures), thereby helping preventing texture baking in $\bf \hat{RM}$ (Fig.~\ref{fig:gm}).
The final loss is the combination of all above losses:
\begin{align}
L_{\theta} = L_\text{MSE}({\bf A}, { \bf \hat A}) + L_\text{MSE}({\bf RM}, {\bf \hat{RM}}) +  L_\text{GM}({\bf RM}, {\bf \hat{RM}}).
\label{equ:final_loss}
\end{align}

\subsection{Detail Refinement}
\label{subsec:detial_refine}
Although optimizing in pixel space can enhance the sharpness of output results, achieving a comparable quality of multi-step denoising in a single step is still challenging. 
Furthermore, the structure information of high-frequency textures is misaligned after passing through the VAE encoder, leading to failures in detail recovery. This has a significant impact on albedo prediction, and also affects the prediction of roughness and metallic properties for objects with high-frequency geometric contours (Fig.~\ref{fig:din_ablation}).

\begin{figure}
\centering
\includegraphics[width=\linewidth]{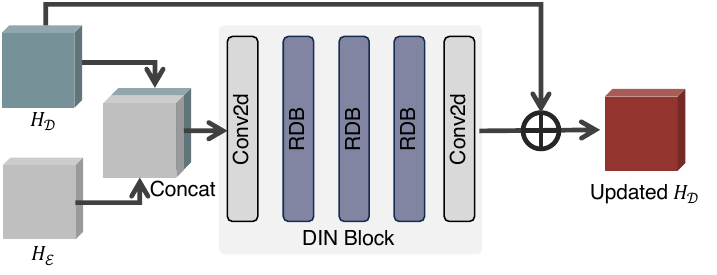}
\caption{\textbf{The architecture of Detail Injection Network (DIN).} }
\label{fig:din} 
\end{figure}

Inspired by~\cite{xu_2025_CVPR,zhu2023designing}, we introduce a detail injection network (DIN) $f_\phi$, which can effectively improve the prediction details by modulating the hidden features of $\mathcal{D}$:
\begin{align}
H_{\mathcal{D}} = f_{\phi}(\text{concat}(H_\mathcal{E},H_\mathcal{D})) + H_{\mathcal{D}},
\label{equ:din}
\end{align}
where $H_\mathcal{E}$ and $H_\mathcal{D}$ are hidden features of encoder $\mathcal{E}$ and decoder $\mathcal{D}$ respectively, and are concatenated along the channel dimension.
Both $H_{\mathcal{E}}$ and $H_{\mathcal{D}}$ are high resolution features, which typically have more detailed information and are suitable for detail injection. As shown in Fig.~\ref{fig:din}, DIN consists of a set of Residual Dense Blocks (RDB)~\cite{zhang2018residual}, which is originally designed for image super-resolution. It takes the concatenation of $H_{\mathcal{E}}$ and $H_{\mathcal{D}}$ as input and outputs a new feature to update $H_{\mathcal{D}}$. 

Although DIN doesn't incorporate cross-view attention to ensure consistency, we didn't observe inconsistency issues in most cases (Fig~.\ref{fig:multi} presents two examples in the multi-view setting). We believe this is because the injected high-frequency structures are naturally multi-view consistent in the original RGB images (e.g., text details are generally consistent across views). Even when there are inconsistent features, such as highlights, reflections, and varying illumination, DIN can still effectively avoid introducing these features. This benefits from being trained on such data and leveraging the denoised multi-view consistent material features $H_\mathcal{D}$ as a reference.

\subsection{Discussion of low-variance}
In our one-step diffusion model, the randomness of inference mainly comes from the initial Gaussian noise. As shown in Fig.~\ref{fig:var}
, compared with multi-step methods, the variance of our model is sufficiently small, making it difficult to visually discern differences in estimated results across different seeds. We think this is because our model is optimized at time step $T$, where the input noise is a pure Gaussian noise providing limited information for material estimation. Therefore, our model shows a low sensitivity to input noise. Consequently, 
even if we remove the Gaussian noise input, it does not adversely affect the model's performance.

\section{Implementation Details}\label{sec:detail}

\paragraph{Network architecture}
Similar to IDArb~\cite{li2024idarb}, we adopt cross-component and cross-view attention to improve prediction quality and enforce global consistency across different views. To switch the task between predicting $\bf A$ and $\bf RM$, we use specific text prompts as conditions. For DIN, we directly use the RDB block from Zhang et al. \shortcite{zhang2018residual} and set the kernel size to 3 for all other convolutional layers. To modulate hidden features through DIN, we use the first two layers of VAE encoder $\mathcal{E}$ to update the last two layers of VAE decoder $\mathcal{D}$.

\paragraph{Optimization}
For better initializing our one-step diffusion model, we first train a multi-step diffusion model for material estimation using $\bf v$-parameterization, with the same training setup as IDArb~\cite{li2024idarb}. Then, we initialize StableIntrinsic with this pre-trained model and optimize $\mu_\theta$ on time step $T$ only. However, simultaneously optimizing $\mu_\theta$ and $f_{\phi}$ is difficult due to the limited GPU memory. 
Therefore, we first use Equ.~\ref{equ:final_loss} to optimize $\mu_\theta$ with a fixed VAE~\cite{kingma2013auto}. Subsequently, we freeze all other network parameters and only optimize $f_{\phi}$ to enhance details.

\paragraph{Hyper-parameters configuration}
For the training of U-net $\mu_{\theta}$ and DIN $f_{\phi}$, we use AdamW~\cite{loshchilov2017decoupled} optimizer with a learning rate of $1 \times 10^{-4}$ that linearly decays to $1 \times 10^{-5}$. $\mu_{\theta}$ is trained over 10K steps with a batch size of 8 and $f_{\phi}$ is trained over 20K steps with a batch size of 4. Both $\mu_{\theta}$ and $f_{\phi}$ are optimized using 32 AMD MI308X GPU cards.

\begin{table}[t]
\centering
\caption{\textbf{Quantitative comparisons on synthetic data}. The best results are highlighted as \textbf{1st} and \underline{2nd}. For real-world datasets, the metrics are calculated by their pseudo ground-truth maps.}
\begin{tabular}{ccccc}
\toprule
\multirow{2}{*}{Methods} & \multicolumn{2}{c}{Albedo} & \multicolumn{1}{c}{Metallic} & \multicolumn{1}{c}{Roughness} \\ 
\cmidrule(lr){2-3} \cmidrule(lr){4-4} \cmidrule(lr){5-5}
& SSIM $\uparrow$ & PSNR $\uparrow$ & MSE $\downarrow$ & MSE $\downarrow$ \\
\midrule
\multicolumn{5}{c}{Objaverse-Testing} \\
\midrule
RGB$\leftrightarrow$X & 0.861 & 25.42 & 0.068 & 0.035 \\
IntrinsicAnything & 0.880 & 26.54 & - & - \\
IDArb & \underline{0.915} & \underline{29.73} & \underline{0.029} & \underline{0.020} \\
Ours & \textbf{0.949} & \textbf{32.67} & \textbf{0.015} & \textbf{0.008} \\
\midrule
\multicolumn{5}{c}{ShinyBlender \cite{verbin2021ref}} \\
\midrule
RGB$\leftrightarrow$X & 0.861 & 22.54 & 0.101 & 0.144 \\
IntrinsicAnything & 0.893 & 26.34 & - & - \\
IDArb & \underline{0.951} & \underline{30.20} & \underline{0.009} & \underline{0.004} \\
Ours & \textbf{0.973} & \textbf{34.28} & \textbf{0.005} & \textbf{0.002} \\
\bottomrule
\end{tabular}
\label{tab:quant_comp_synth}
\end{table}

\begin{table}[t]
\centering
\caption{\textbf{Quantitative comparisons on real-world data}. We compare our \textbf{albedo} results against other baselines. The best results are highlighted as \textbf{1st}, \underline{2nd}. The metrics are calculated using the pseudo ground-truth maps.}
\begin{tabular}{ccccc}
\toprule
\multirow{2}{*}{Methods} & \multicolumn{2}{c}{Stanford-ORB} & \multicolumn{2}{c}{MIT-Intrinsic} \\ 
\cmidrule(lr){2-3} \cmidrule(lr){4-5}
& SSIM $\uparrow$ & PSNR $\uparrow$ & SSIM $\uparrow$ & PSNR $\uparrow$ \\
\midrule
RGB$\leftrightarrow$X & 0.839 & 30.23 & 0.793 & 22.43 \\
IntrinsicAnything & 0.847 & 31.83 & \underline{0.822} & \underline{24.42} \\
IDArb & \underline{0.866} & \textbf{32.57} & 0.811 & 22.84 \\
Ours & \textbf{0.882} & \underline{32.33} & \textbf{0.842} & \textbf{25.72} \\
\bottomrule
\end{tabular}
\label{tab:quant_comp_real}
\end{table}

\section{Experiments}\label{sec:exp}

\subsection{Setup}\label{subsec:setup}

\paragraph{Dataset}
For training, we use the same training set as IDArb \cite{li2024idarb}, which employs Arb-Objaverse \cite{li2024idarb}, G-Objaverse \cite{zuo2024sparse3d}, and ABO \cite{collins2022abo} with a resolution of $256\times 256$.
For evaluation, since the testing split of IDArb is unpublished, we exclude object indices present in the training set and select the remaining objects from Arb-Objaverse along with a random subset from G-Objaverse, forming the Objaverse-Testing dataset of 200 objects. To demonstrate the generalizability of our method, we also evaluate on ShinyBlender \cite{verbin2021ref} and two public real-world datasets: MIT-Intrinsic \cite{grosse2009ground} and Stanford-ORB \cite{kuang2023stanfordorb}.
We sample 4 viewpoints for each object from multi-view datasets with a resolution of $512\times 512$ for evaluation. As the background occupies a large portion that can affect quantitative comparisons, we crop each input image based on its mask to enhance the precision of metric calculations. 

\paragraph{Baselines}
To evaluate our method, we conduct comparisons with the state-of-the-art diffusion-based material estimation methods: RGB$\leftrightarrow$X~\cite{zeng2024rgb}, IntrinsicAnything~\cite{chen2024intrinsicanything}, and IDArb~\cite{li2024idarb}. We utilize their pre-released models for evaluation. 
As IntrinsicAnything only releases the single-image inference code and does not decompose the metallicity and roughness, we adopt it for albedo comparison in the single-view setting.Note that RGB$\leftrightarrow$X is trained on scene data, yet it demonstrates generalizability when applied to object-centric data. Therefore, we include it as a baseline in our comparisons.

\paragraph{Metrics}
Similar to IDArb~\cite{li2024idarb}, we use a scale-invariant Peak Signal-to-Noise Ratio (PSNR) and Structural Similarity Index Measure (SSIM)~\cite{zhou2004ssim} for evaluating albedo, and Mean Square Error (MSE) for evaluating metallicity and roughness.

\begin{figure}[t]
\centering
\includegraphics[width=\linewidth]{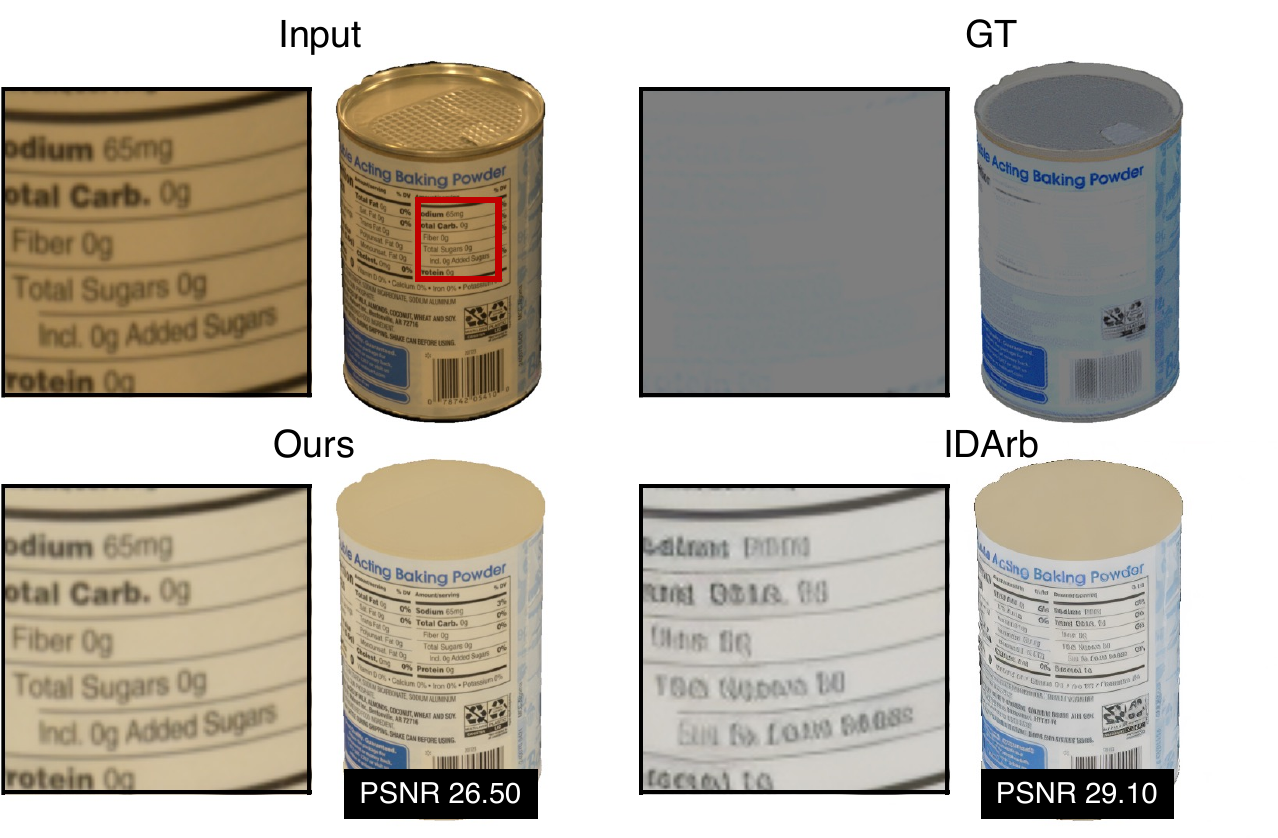}
\caption{\textbf{Pseudo ground-truth albedo maps of Stanford-ORB~\cite{kuang2023stanfordorb}.}
Since the ground truth of the Stanford-ORB dataset is pseudo, our method might achieve a lower PSNR compared to other methods, even though it produces sharper albedo maps.}
\label{fig:pse_orb} 
\end{figure}

\begin{table}[t]
\centering
\caption{\textbf{Ablation Studies.} 
w/o Opt. represents that we directly use the pre-trained $\bf v$-parameterization diffusion model to conduct one-step denoising. In order to eliminate the influence of DIN, the following ablations are conducted \textbf {without using DIN}, except for the last row (Ours), which is used to show the effect of DIN.}
\begin{tabular}{lcccc}
\toprule
& \multicolumn{2}{c}{Albedo} & \multicolumn{1}{c}{Metallic} & \multicolumn{1}{c}{Roughness} \\ 
\cmidrule(lr){2-3} \cmidrule(lr){4-4} \cmidrule(lr){5-5}
& SSIM $\uparrow$ & PSNR $\uparrow$ & MSE $\downarrow$ & MSE $\downarrow$ \\
\midrule

w/o Opt. & 0.919 & 30.40 & 0.019 & 0.011 \\
w/o $L_{\rm GM}$ & 0.929 & 30.83 & 0.018 & 0.011 \\
Ours (w/o DIN) & 0.932 & 31.33 & 0.016 & 0.009 \\
Ours (w/ DIN) & \textbf{0.949} & \textbf{32.67} & \textbf{0.015} & \textbf{0.008} \\

\bottomrule
\end{tabular}
\label{tab:ablation}
\end{table}

\subsection{Comparison}\label{subsec:comp}

\begin{figure*}[t]
\centering
\includegraphics[width=\linewidth]{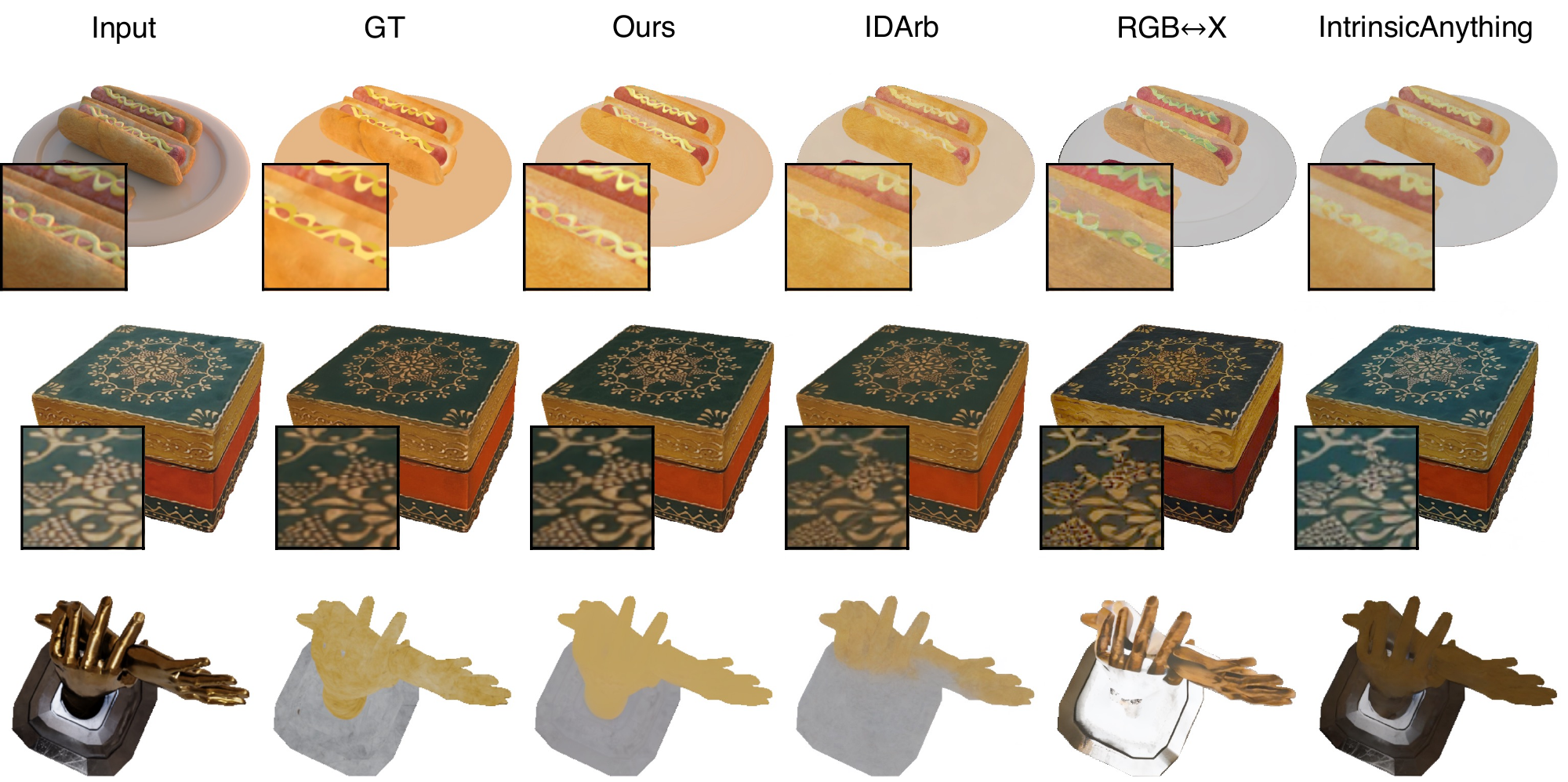}%
\caption{\textbf{Albedo qualitative comparison on synthetic data TensoIR~\cite{Jin2023TensoIR} (top) and Objaverse-Testing (middle and bottom).} Our method successfully preserves accurate textures and reliably distinguishes reflections, resulting in the best albedo estimation.}
\label{fig:albedo_com_syn} 
\end{figure*}

\begin{figure*}[t]
\centering
\includegraphics[width=0.93\linewidth]{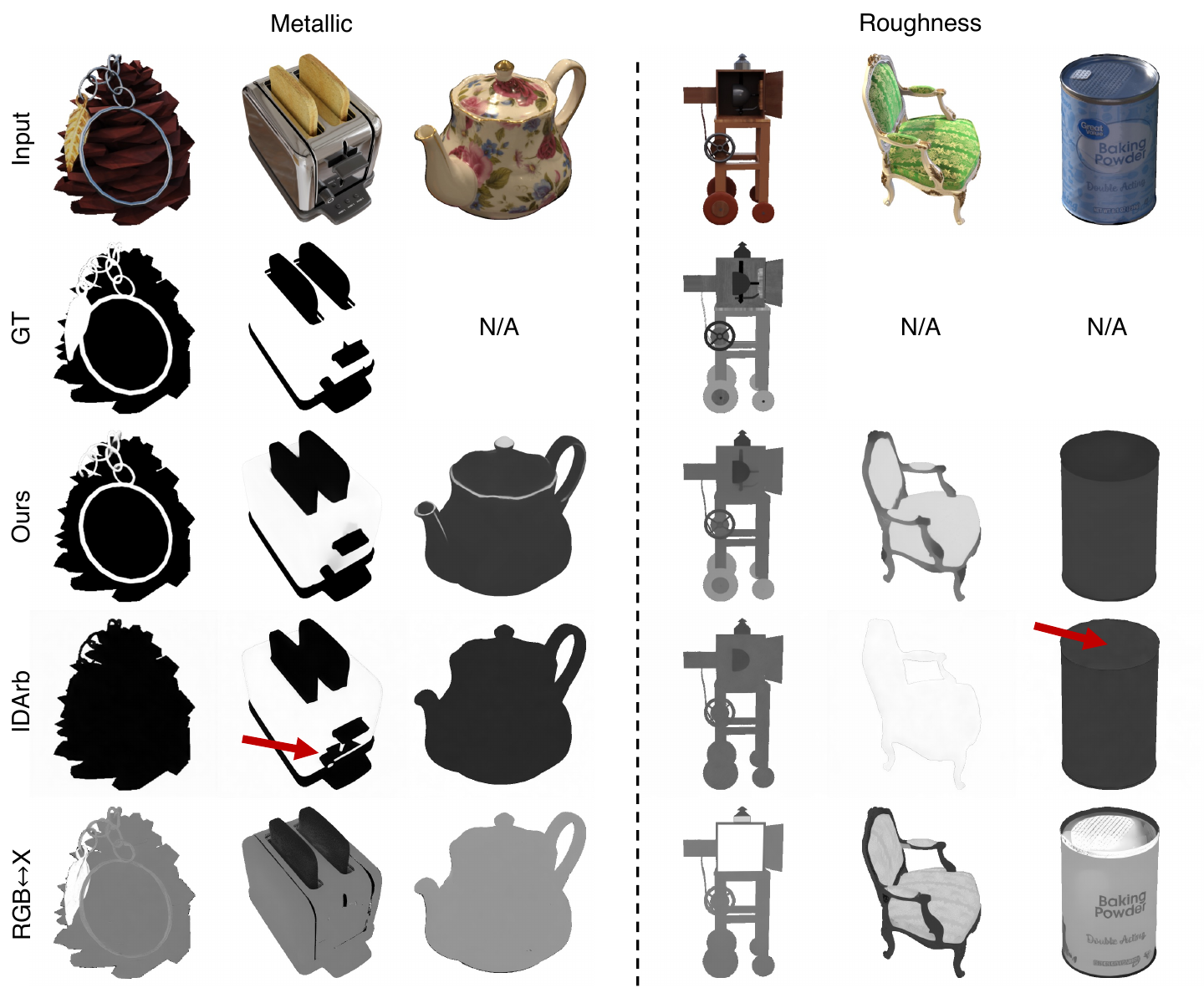}%
\caption{\textbf{Metallicity and roughness comparison on both synthetic data and real-world data.} We use \textit{N/A} to indicate that no ground-truth is available for this data. While IDArb \cite{li2024idarb} has the limitation of oversimplification, our method stably eliminates the influence of shadows and reflections, and produces more plausible results for parts with varying material properties.}
\label{fig:rm_comp} 
\end{figure*}

A quantitative comparison on synthetic data is presented in Tab.~\ref{tab:quant_comp_synth}, where our method achieves superior results across all datasets. 
For Objaverse-Testing,
our model produces the best results, demonstrating its capability to effectively handle various types of objects. 
Two examples are shown in Fig.~\ref{fig:albedo_com_syn}, where our model recovers the specific patterns and distinguishes reflections.
For ShinyBlender, we re-rendered the images and ground-truth material maps from the original project files. Despite the minimal difference in MSE metrics between our model and IDArb, our results successfully recover metallic properties without baked reflections, as displayed in Fig.~\ref{fig:rm_comp} in the second column. This improvement will greatly benefit downstream operations, such as inverse rendering.

\begin{figure*}[t]
\centering
\includegraphics[width=\linewidth]{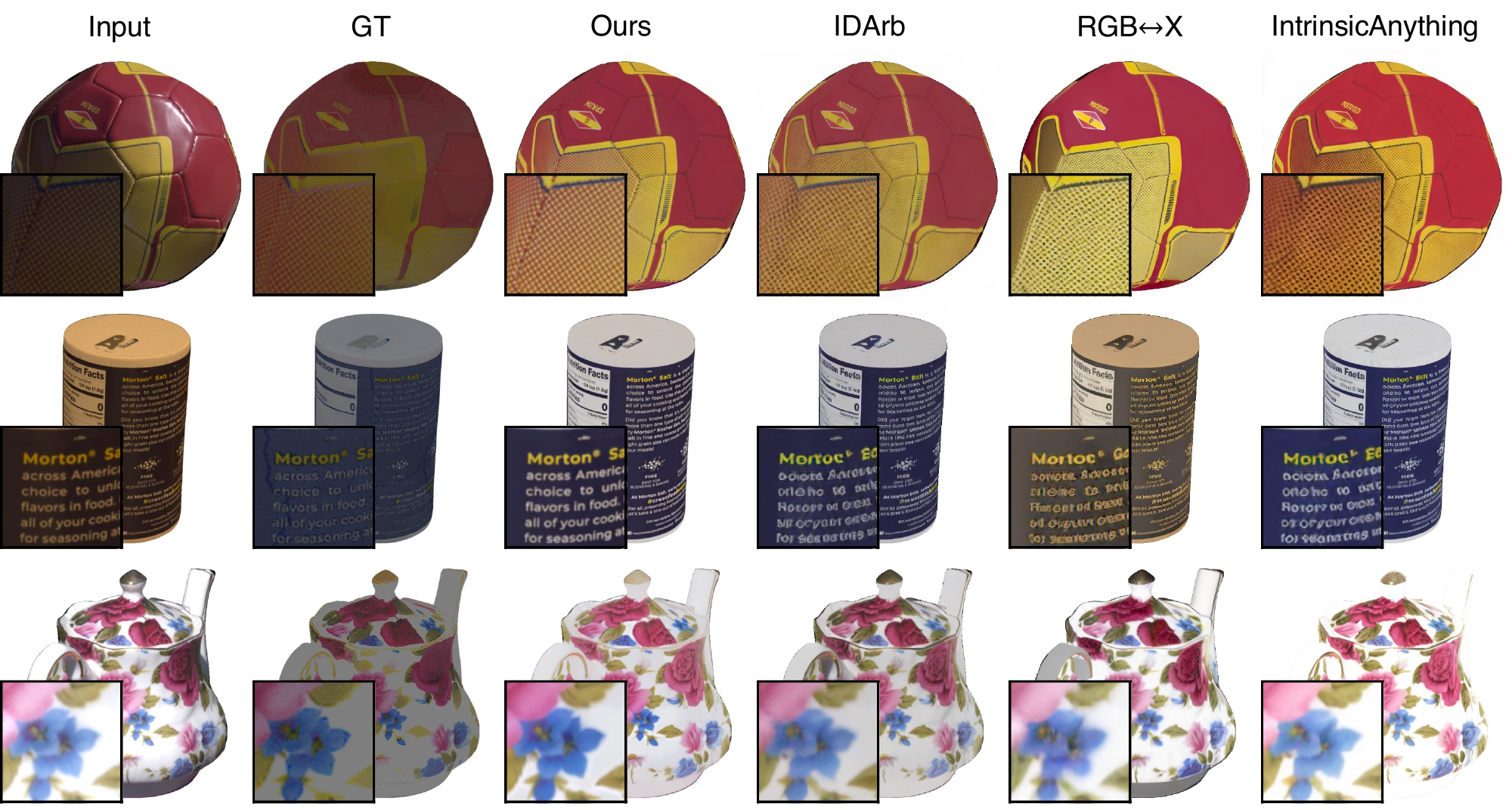}%
\caption{\textbf{Albedo qualitative comparison on public real-world data Standford-ORB~\cite{kuang2023stanfordorb}.} Note that the \textit{GT} maps are pseudo ground-truth albedo maps obtained from the original works, and are displayed as a reference. Our method recovers fine-grained structured textures and effectively removes highlights and shadows.}
\label{fig:albedo_com_real} 
\end{figure*}

For real-world data, we calculate the metrics on albedo using the pseudo ground-truth maps, as shown in Tab.~\ref{tab:quant_comp_real}.
Our method performs slightly worse in PSNR compared to IDArb on the Stanford-ORB dataset.
However, it is noteworthy that the pseudo albedo maps provided by Stanford-ORB often lack details (as exemplified in Fig.~\ref{fig:pse_orb}).
Furthermore, as the images are captured under environmental illumination with varying colors, decomposing color from albedo becomes highly ambiguous. Our results demonstrate a more consistent illumination decomposition across the entire object, although this does not translate into higher metric scores.
We display more qualitative results in Fig.~\ref{fig:albedo_com_real} and ~\ref{fig:rm_comp}.
While other approaches struggle with the loss of structured details and the stochastic generation of high-frequency textures, our model 
recovers the fine-grained textures and more plausible material properties.

Tab.~\ref{tab:efficiency_parallel} shows the inference efficiency comparison for both single-view and multi-view settings, tested on a single NVIDIA Tesla V100 GPU card. Since both IDArb and RGB$\leftrightarrow$X require 50 denoising steps (default setting), our method achieves a denoising speed approximately 50 times faster than IDArb and 30 times faster than RGB$\leftrightarrow$X (Note that RGB$\leftrightarrow$X does not incorporate cross-view/cross-component attention, making its denoising slightly faster than IDArb, but it lacks multi-view consistency.). Additionally, the introduction of DIN results in our decoding being slightly slower than that of the other methods. Overall, our model is significantly more efficient than the others.

\begin{table}[t]
\centering
\caption{\textbf{Inference efficiency}. We show the time cost (in seconds) of each component during inference. The best results are highlighted as \textbf{1st}.}
\resizebox{\linewidth}{!}{
\begin{tabular}{l|ccc|ccc}
\toprule
\multirow{2}{*}{} & \multicolumn{3}{c|}{Single-view} & \multicolumn{3}{c}{Multi-view} \\
& Our & IDArb & RGB$\leftrightarrow$X & Our & IDArb & RGB$\leftrightarrow$X \\
\midrule
Encode & 0.046 & 0.046 & 0.032 & 0.190 & 0.165 & 0.106 \\
Denoise & 0.094 & 5.781 & 3.307 & 0.402 & 26.019 & 10.665 \\
Decode & 0.441 & 0.303 & 0.231 & 1.752 & 1.155 & 0.899 \\
\midrule
Total & \textbf{0.581} & 6.130 & 3.570 & \textbf{2.344} & 27.339 & 11.670 \\
\bottomrule
\end{tabular}}
\label{tab:efficiency_parallel}
\end{table}

\subsection{Ablation Studies}
\paragraph{Detail Inject Network} 
We disable the DIN to evaluate its impact. For albedo, as shown in Fig.~\ref{fig:din_ablation}, DIN can effectively improve the details and correct the structure information, making the letters on the can as clear as the input condition image. For roughness and metallic, DIN improves them by enhancing the sharpness of boundaries (Fig.~\ref{fig:din_ablation} bottom). Since albedo contains more texture information, the improvement effect of DIN on albedo is more significant compared to roughness and metallic (Tab.~\ref{tab:ablation}). 

\begin{figure}[t]
\centering
\includegraphics[width=\linewidth]{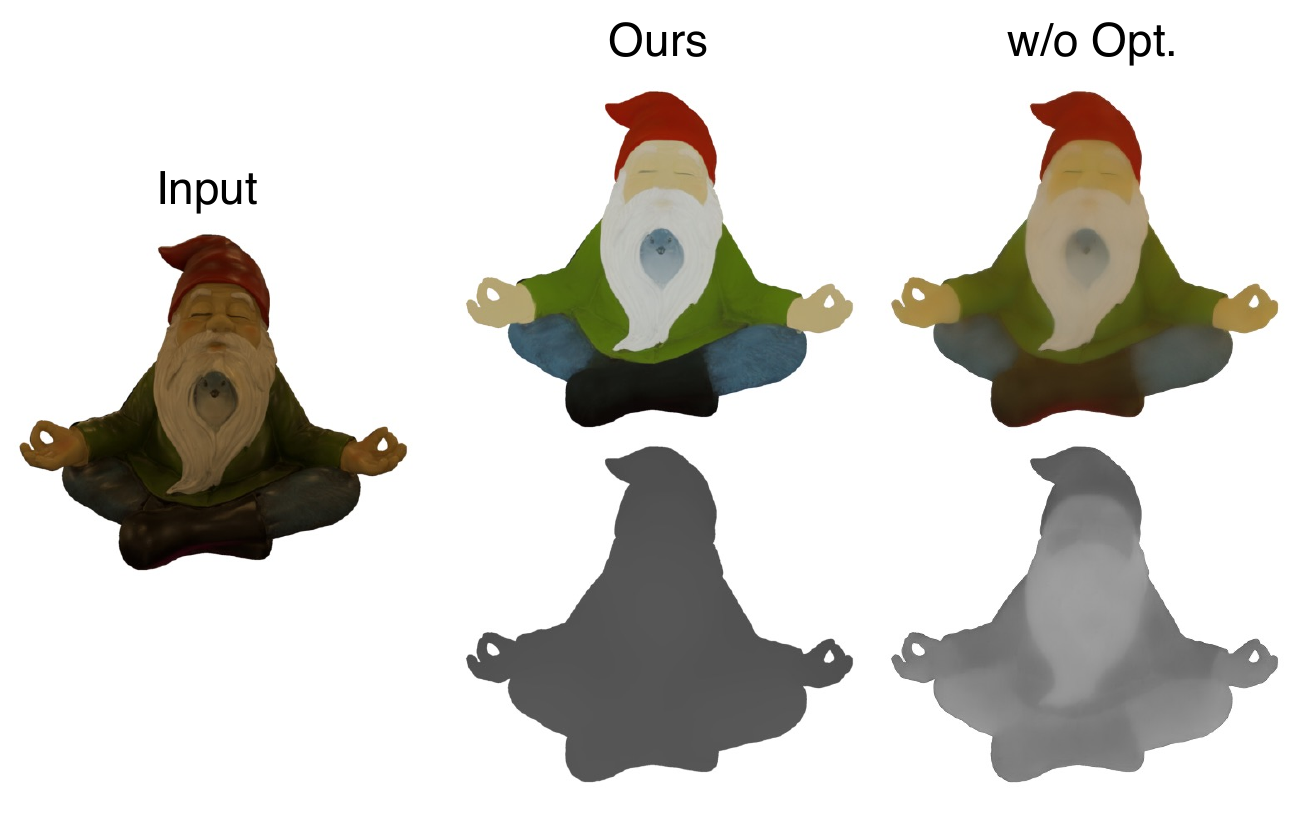}
\caption{\textbf{StableIntrinsic improves the results of one-step diffusion model}. \textbf{Top:} estimated albedo maps; \textbf{Bottom:} estimated roughness maps.}
\label{fig:opt} 
\end{figure}

\begin{figure}[t]
\centering
\includegraphics[width=\linewidth]{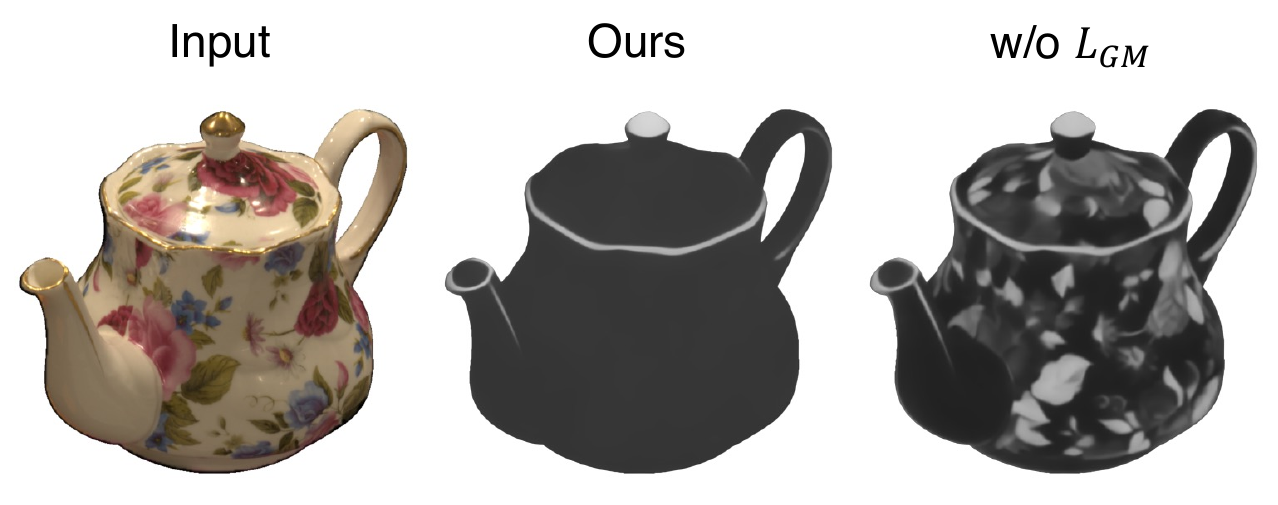}
\caption{\textbf{Ablation of loss $L_{\rm GM}$}. Applying $L_{\rm GM}$ makes the predicted metallic of teapot more accurate.}
\label{fig:gm} 
\end{figure}

\paragraph{Optimization}
To demonstrate the necessity of further optimization on $t=T$, we compare our results with one-step denoising results of the pre-trained diffusion model (Tab.~\ref{tab:ablation}). 
We set the number of DDIM steps to 1 ($t = T$) for the pre-trained model to obtain the one-step denoising results.
Fig.~\ref{fig:opt} shows that, without optimization, one-step denoising produces blurry and incorrect results, while our model estimates sharper albedo and more accurate roughness.

\paragraph{Gradient matching loss}
The gradient matching loss $L_{\rm GM}$ prevents textures from being baked into the predicted roughness and metallic maps (Fig.~\ref{fig:gm}). Although $L_{\rm GM}$ is only applied to roughness and metallic, removing $L_{\rm GM}$ leads to decreased PSNR and SSIM scores of the predicted albedo as well (Tab.~\ref{tab:ablation}).
This might be because
a lower quality of roughness or metallic influences the estimation of albedo through cross-component attention. 

\begin{figure*}[t]
\centering
\includegraphics[width=\linewidth]{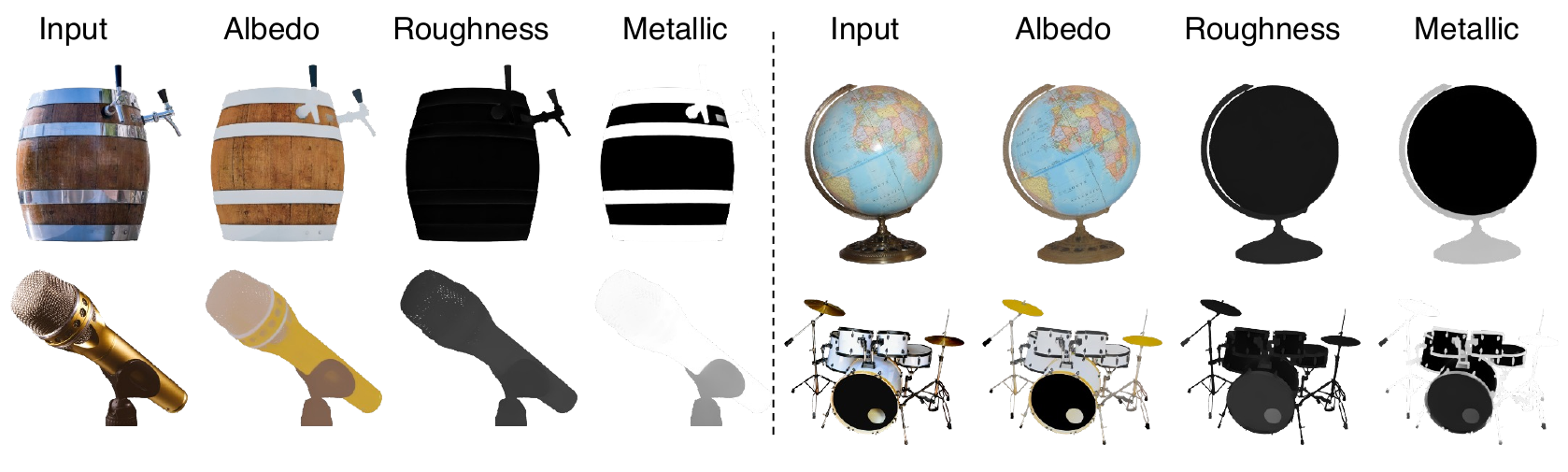}%
\caption{\textbf{Results on real data.}}
\label{fig:real_data} 
\end{figure*}

\begin{figure*}[t]
\centering
\includegraphics[width=\linewidth]{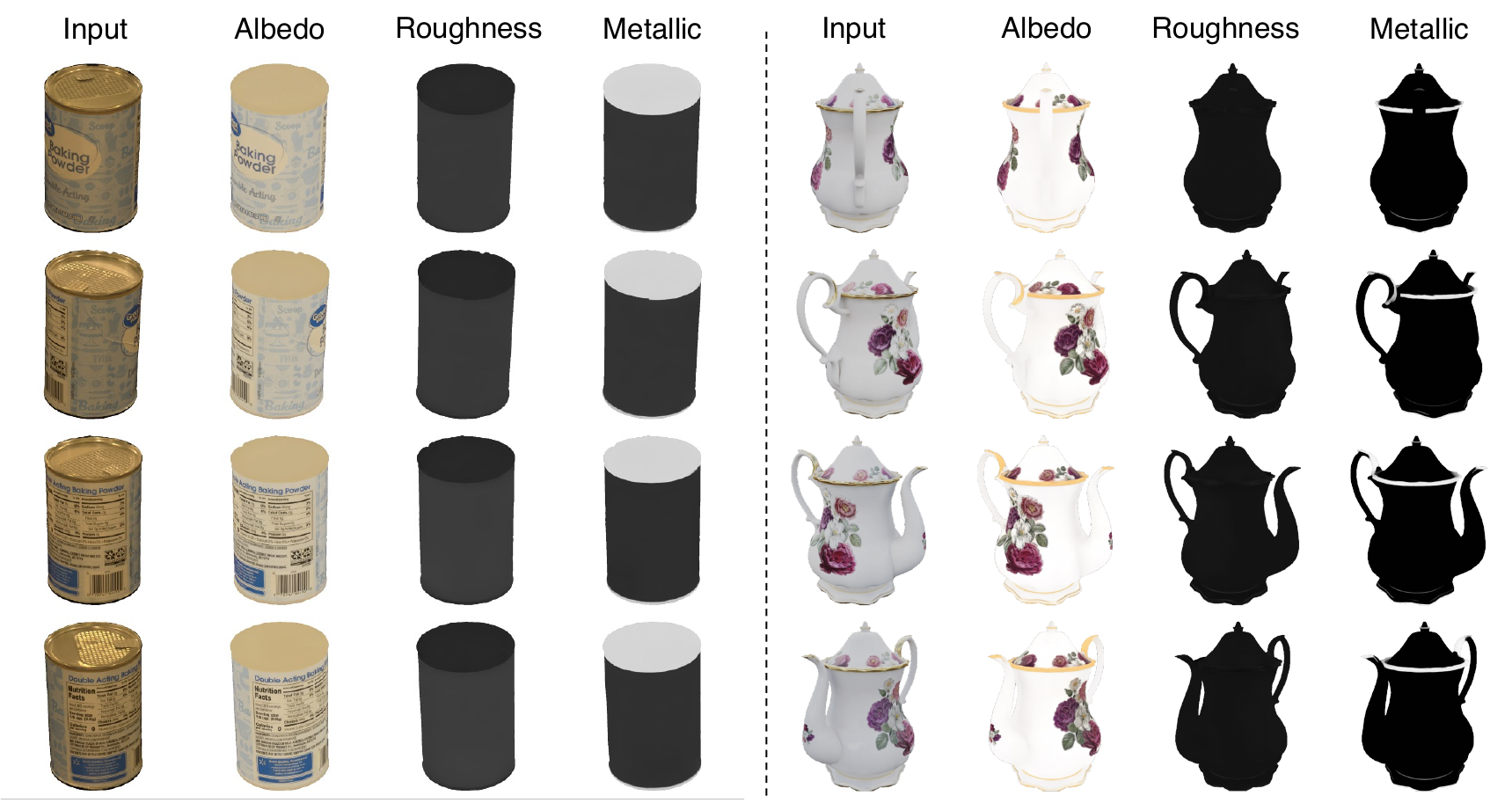}%
\caption{\textbf{Multi-view estimated results.}}
\label{fig:multi} 
\end{figure*}

\paragraph{More results}
We conduct extra experiments to fully show the capability of the model.
Fig.~\ref{fig:real_data} presents more results on real data collected from the Internet\footnote{https://pixabay.com/}, and Fig.~\ref{fig:multi} demonstrates the consistency across multiple views.

\begin{figure*}[t]
\centering
\includegraphics[width=\linewidth]{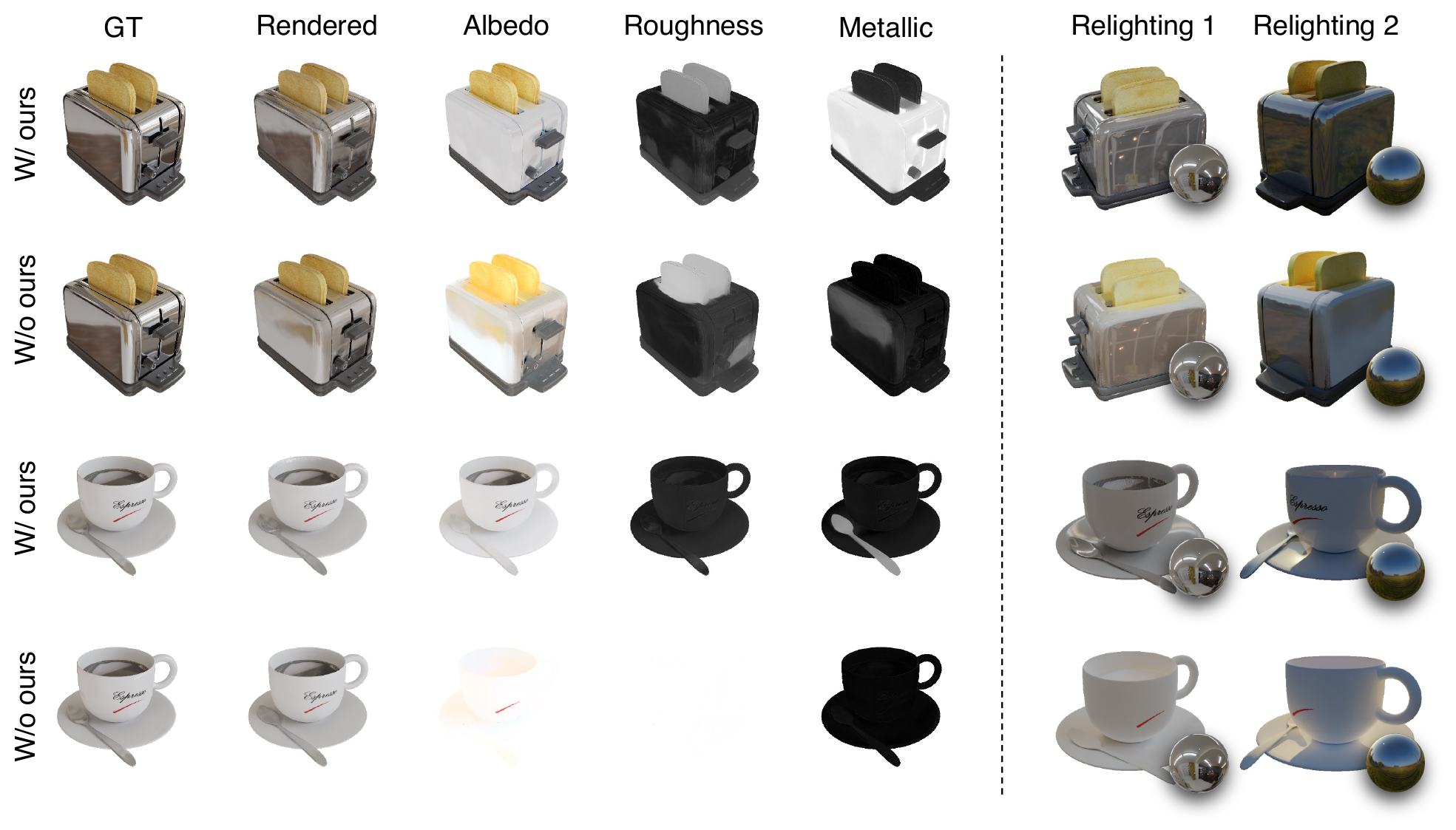}%
\vspace{-0.5cm}
\caption{\textbf{Applications.} We show the reconstructed results of NVDiffRec~\cite{munkberg2022extracting} both with and without supervision from our model's predictions.}.
\label{fig:application} 
\end{figure*}

Furthermore, the predicted results can be applied to enhance the material decomposition for inverse rendering. Specifically, we leverage the predictions of our model to supervise NVDiffRec~\cite{munkberg2022extracting} during the optimization of albedo and RM maps by applying an MSE loss. As shown in Fig.~\ref{fig:application}, the inverse rendering framework, guided by our priors, achieves a more plausible decomposition of materials and illumination, demonstrating the practical value of StableIntrinsic for downstream applications.

\section{Conclusion}\label{sec:conclusion}
\begin{figure}[h]
\centering
\includegraphics[width=\linewidth]{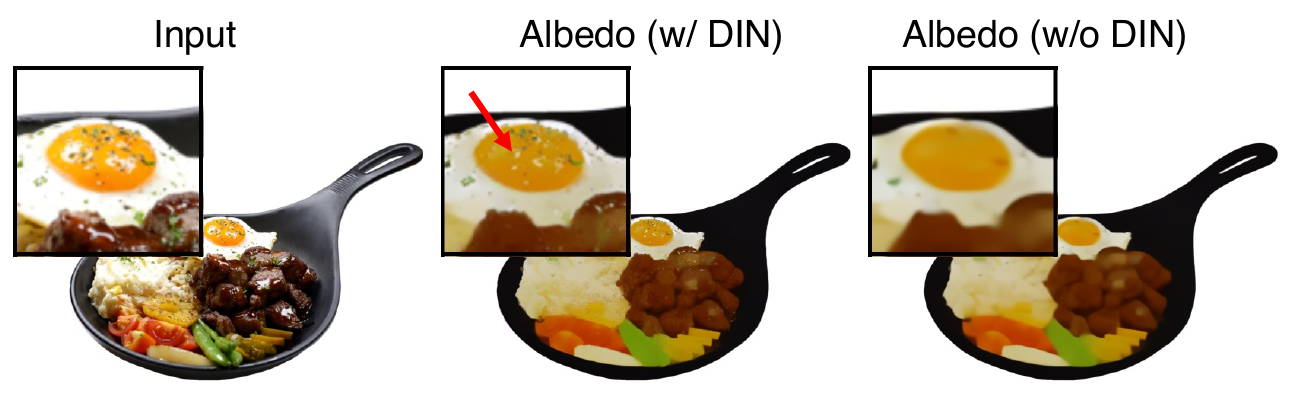}
\caption{\textbf{Failure case}. 
DIN incorrectly injects highlights in areas with details.}
\label{fig:fc} 
\end{figure}

We have demonstrated a one-step diffusion model, StableIntrinsic, which estimates material parameters from multi-view RGB images while highly preserving details. However,
the estimated results of our model might be affected by 
the illumination conditions, showing different output results for same objects under different illuminations. While this issue does not occur within a single inference batch, it restricts the application to reconstructing material maps from a plethora of images that require to be processed in multiple batches. The source of this problem mainly comes from the limited diversity of illumination in the training dataset. 
In the future, the model's capability can be enhanced by expanding the training data and increasing the diversity of lighting conditions.
In addition, although our DIN can effectively inject details into the predicted results, it might incorrectly introduce some high-frequency highlights into the albedo maps. This usually occurs on objects with strong lighting and rich textures, primarily because the network struggles to distinguish between small sharp highlights and texture details (see Fig.~\ref{fig:fc} for an example). Therefore, it is highly valuable to develop a DIN with stronger capabilities in highlight awareness.

\clearpage
\clearpage

\bibliographystyle{ACM-Reference-Format}
\bibliography{reference}

\end{document}